# Discovering influential text using convolutional neural networks

**Megan Ayers**[1*], **Luke Sanford**[1*], **Margaret E. Roberts**[2], **Eddie Yang**[2]
[1]Yale University    [2]University of California San Diego
{m.ayers,luke.sanford}@yale.edu, {meroberts,z5yang}@ucsd.edu

## Abstract

Experimental methods for estimating the impacts of text on human evaluation have been widely used in the social sciences. However, researchers in experimental settings are usually limited to testing a small number of pre-specified text treatments. While efforts to mine unstructured texts for features that causally affect outcomes have been ongoing in recent years, these models have primarily focused on the topics or specific words of text, which may not always be the mechanism of the effect. We connect these efforts with NLP interpretability techniques and present a method for flexibly discovering clusters of similar text phrases that are predictive of human reactions to texts using convolutional neural networks. When used in an experimental setting, this method can identify text treatments and their effects under certain assumptions. We apply the method to two datasets. The first enables direct validation of the model's ability to detect phrases known to cause the outcome. The second demonstrates its ability to flexibly discover text treatments with varying textual structures. In both cases, the model learns a greater variety of text treatments compared to benchmark methods, and these text features quantitatively meet or exceed the ability of benchmark methods to predict the outcome.

## 1 Introduction

Text impacts outcomes and decisions in many domains. For example, researchers have investigated the effects of campaign messaging on voting (Arceneaux and Nickerson, 2010), post content on censorship (King et al., 2014), clinical notes on diagnoses and treatment (Sheikhalishahi et al., 2019), and written profiles on citizenship decisions (Hainmueller and Hangartner, 2013). Most experimental methods for estimating the effects of text on human evaluation randomly assign some subjects to a small number of treatment texts which are chosen ex-ante by the researcher. These treatments are often chosen subjectively, introducing the possibility that they may be ineffective or lack external validity. Recent literature in computational social science has sought to instead discover treatments that have an effect on an outcome of interest from unstructured texts (Fong and Grimmer, 2016; Pryzant et al., 2018). In this work, we bridge these efforts towards making causal inferences using text treatments with the domain of interpretable machine learning (Jacovi et al., 2018; Alvarez Melis and Jaakkola, 2018). Causal inference approaches with text as treatment aim to identify low-dimensional representations of text features that causally impact an outcome. We introduce an innovative application of contextualized word embeddings, convolutional neural networks (CNNs), and interpretability methods to this setting to both detect and interpret these latent text representations. Unlike previous approaches to identifying text treatments, these learned representations can vary in both length and structure and are not constrained to represent document-level sets of topics or sets of particular words. We apply our approach to two datasets: social media posts on Weibo where the outcome is post censorship, and complaints submitted to the Consumer Financial Protection Bureau with the outcome of whether a complainant received a timely response. In both cases, our method extracts qualitatively different treatments and meets or exceeds quantitative performance metrics compared to benchmark methods.

While this approach and those that it builds upon seek to simulate experiments that tar-get causal effects of specific latent text features, effects can only be identified directly under strong assumptions. Alternatively, this model may be used as a tool for researchers to discover text treatments to test in subsequent confirmatory analyses, as an alternative to subjectively posing treatments.

---
[*]These authors contributed equally to this work.

## 2 Related work

While much of the related social science work has focused on learning latent "features" of texts and using those as a treatment, most NLP work has focused on improving the interpretability of black-box predictive models. This paper bridges the two by using interpretability methods to flexibly discover latent treatments in text.

**Computational social science and causal inference** Prior work has generated methods to both discover treatments and estimate their effects simultaneously (Fong and Grimmer, 2016; Pryzant et al., 2018; Egami et al., 2018; Fong and Grimmer, 2021; Feder et al., 2022). These models have typically focused on estimating either topics or individual words as treatments. Fong and Grimmer (2016) apply a supervised Indian buffet process to both discover features (topics) and estimate their effect on an outcome in an RCT setting. Pryzant et al. (2018) use n-gram features instead of topics and a neural architecture with a method for extracting feature importance from the weights of the network. Our model extends this work by allowing groups of generally similar phrases – instead of topics or unique words – to be identified as treatments. We expect our approach to work particularly well in instances where the outcome may be caused by flexibly expressed concepts (e.g. a sentiment that can be conveyed with interchangeable synonyms) instead of particular words or the full topical content of the document.

**Interpretable NLP** Many methods have been proposed to interpret and explain NLP models, as well as meta-evaluations of those methods (Lei et al., 2016; Alvarez Melis and Jaakkola, 2018; Rajagopal et al., 2021; Alangari et al., 2023; Crothers et al., 2023; Lyu et al., 2023). Most of these methods focus on explaining and interpreting predictions at the level of individual samples. In contrast, our method is designed to learn and interpret broader patterns that occur at the corpus level. In this respect, Rajagopal et al. (2021), who require their model to explain predictions using "global" concepts, and Jacovi et al. (2018), who interpret the latent features learned by CNNs specifically, are closest to our work. Individual tokens are not human-interpretable or individually persuasive, so like Alvarez Melis and Jaakkola (2018) we require the network to have an interpretable final layer after a representation learning component. Rather than trying to understand why the network made the prediction it did, we seek representations of influential corpus-wide features whose effects scientists can test in follow-up experiments. For example, if the model identifies the presence of calendar dates as a globally influential feature for determining timely response to a complaint, researchers may design two texts differing only by the inclusion of dates and compare their effects in a controlled experiment.

Other existing NLP techniques could be adapted to this approach. For example, the differences between persuasive and unpersuasive texts (Zhong et al., 2022) could be used to identify persuasive concepts. While any method capable of learning corpus-wide and low-dimensional representations of influential text features could be utilized for identifying text treatments, a key challenge is the ability to capture complex feature representations while remaining human interpretable. This requires sophistication in representation learning but clarity in understanding the learned text treatments whose effects are being estimated. In experiments, our proposed model effectively achieves this balance.

## 3 Extracting influential text from latent representations

Our goal is to extract clusters of phrases that represent latent, generalizable treatments that affect a particular outcome. To do this, we imagine that $N$ texts ($T_i$) are randomly assigned to a process through which they are mapped to an outcome ($Y_i$). Let $i$ also index the individual evaluating text $i$. We seek to identify and estimate the effect of an $m$-dimensional latent representation of those texts ($Z_i$) which summarizes clusters of phrases or concepts that are likely to influence the outcome in repeated experiments. We refer to $Z_i$ as "text treatments" for text $i$. For example, each element of $Z_i$ could represent the presence or absence of a certain phrase or grammatical structure, with $Z_i \in \{0,1\}^m$. $Z_i$ could also contain real-valued elements indicating continuous text features like similarity to a certain vocabulary or concept alignment.

To simulate a sequential experimental setup, we follow Egami et al. (2018) in splitting our sample into training and test sets. We first train our model, using cross-validation within the training set for tuning and model selection. We then use the test data set to interpret the latent text treatments discovered and estimate their effects on the outcome under additional assumptions. Our main contribution concerns this first stage: the novel usage of a CNN model to discover a mapping between text data and text treatments ($Z_i$).

Fong and Grimmer (2016, 2021) outline conditions under which this general process identifies causal effects of the text treatments on the outcome when treatments are binary. They assume that: 1) an individual's treatment depends only on their assigned text, 2) any non-textual features or latent text features not captured by the model which influence the evaluator's response ($Y_i$) are independent from the model's captured latent features, 3) there is a nonzero probability of each evaluator receiving any of the possible text treatments ($Z_i$), given unmeasured text features[1], 4) texts are randomly assigned and 5) latent treatments are not perfectly collinear. If these assumptions hold, our model can identify treatment effects of the discovered latent features. These may be estimated using linear regression under the additional assumption that the $m$ text treatments do not interact with each other, in addition to linear modeling assumptions in the case of continuous treatment variables.[2] However, since it is difficult to assess whether these assumptions hold – particularly assumption 2 – we recommend that when possible, practitioners use our method to suggest treatments for study in a controlled experimental setting.

## 4 Methodology

We propose harnessing the structure of CNNs to identify influential text treatments. Filters in convolutional layers project text phrases onto lower-dimensional representation spaces, and these representations are then max-pooled across all phrases within each sample to predict an outcome (Figure 1). By training the model to produce predictive max-pooled representations, filters are incentivized to detect influential n-gram patterns (Jacovi et al., 2018). These patterns could correspond to specific keywords or clusters of keywords with similar vocabulary, grammatical structure, or tone, for example. Researchers can then test how the presence of these patterns in texts affects the outcome.

### 4.1 Contextual encoder

We use pre-trained BERT models (Devlin et al., 2019) to tokenize our input text samples ($T_i$) and to obtain context-dependent embeddings of tokens. We denote these embeddings by $e_{i,j} \in \mathbb{R}^D$, where $i$ indexes each text sample, $j$ indexes tokens ($u_{i,j}$), and $D$ represents the embedding dimension. With accessibility for social scientists in mind, we work with reduced-size models (Jiao et al., 2020), and do not perform fine-tuning. Researchers with fewer constraints on their computational budgets may find improved model performance from using larger or more complex models and/or fine-tuning these models on their outcome. Any model providing text embeddings could be substituted for BERT. However, we do recommend using models that encode context between tokens. We perform the embedding step just before creating a train-test split, but researchers who choose to fine-tune their embedding models should reverse these steps to fine-tune and train only on the training set.

---

[1] For real-valued treatments, this assumption should be modified to require that the probability density function of the treatment vector is nonzero.

[2] Fong and Grimmer (2016) consider the Average Marginal Component Specific Effect (AMCE), which captures the effect of changing one text treatment while averaging over values of all others. For continuous treatments, the process would identify a similar effect capturing the marginal effect of incrementally increasing a text treatment.

## 4.2 Model architecture

Sequences of input text embeddings $\{e_{i,j}\}_j$ are passed to a one dimensional convolutional layer $C$, or a series of $M$ such layers in parallel ($C_l$), each with flexible kernel size $K_l$ and $F$ filters. The number of parallel convolutional layers is determined by the number of unique kernel sizes to be considered. A higher number of filters $F$ corresponds to learning more latent text features. In our implementation all convolutional layers learn the same number of filters. The kernel size $K$ determines the size of the filter window, or the length of phrases considered by each convolutional layer. Including filters of multiple kernel sizes allows the model to capture patterns of varying lengths. A filter $f$ in a layer $C$ with $K = 5$ tests the extent to which the representation learned by $f$ is present in five-token phrases of the input text. For each phrase $p_{i,1}, \ldots, p_{i,P} \in \mathbb{R}^{K \times D}$ in text $i$ with $P = U - K + 1$ and filter $f$, the convolutional operation produces a new feature $a_{i,f} = g(W_f \cdot p_i + b)$, where $W_f$ and $b$ are the learned weights and bias respectively for filter $f$, and $g$ is the sigmoid activation function. We refer to these features as "filter activations," $a_{i,f} \in \mathbb{R}^P$. These are summarized per text sample by max pooling layers, which keep only the highest activation across a text's phrases per filter. The max-pooled activations $a_{i,f}^{pooled} \in \mathbb{R}$ for each filter are then concatenated across the parallel convolutional layers. The concatenated max-pooled activations are then passed through a final fully connected layer. The activations from this final layer, $\hat{Y}_i$, correspond to the model predictions.

## 4.3 Training

The model is trained using Adam optimizer (Kingma and Ba, 2017) and the following loss function when $M = 1$:

$$\mathcal{L} = -\frac{1}{N} \sum_i \left( Y_i \log(\hat{Y}_i) + (1 - Y_i) \log(1 - \hat{Y}_i) \right) +$$

$$\lambda_{ker}^{conv} \sum_{k,d,f} (W_{k,d,f}^{conv})^2 + \lambda_{act}^{conv} \max(R) + \lambda_{ker}^{out} \sum_f |W_f^{out}|$$

where $R$ is an $F \times F$ matrix with

$$R_{f,g} = \begin{cases} \max(\text{cor}(\tilde{a}_f, \tilde{a}_g), 0) & \text{for } f \neq g \\ 0 & \text{for } f = g \end{cases}$$

and $\tilde{a}_f \in \mathbb{R}^{N \cdot P}$ represents the vector of filter activations across $P$ phrases of all $N$ samples for filter $f$. The first term is the binary cross-entropy loss with respect to the model predictions across samples $i$. This global loss was chosen because of the binary outcome in both applications presented here, but could easily be substituted for RMSE or another loss more suitable for continuous outcomes. The second term corresponds to a L2 regularization penalty applied to the weights of the convolutional layers, $W^{conv} \in \mathbb{R}^{K \times D \times F}$. The third term represents an activity regularization which penalizes the maximum non-negative correlation between two filter activations. This penalizes models that learn redundant filters (as measured by high correlation in their activations) to encourage convolutional layers to identify a larger number of distinct text features (Appendix A: Figure 4). For models with $M > 1$, terms two and three are repeated in the sum for each convolutional layer. The fourth term corresponds to a L1 regularization penalty applied to the weights of the final fully connected layer, $W^{out} \in \mathbb{R}^{F \cdot M}$. The strength of each penalty is controlled by $\lambda_{ker}^{conv}$, $\lambda_{act}^{conv}$, and $\lambda_{ker}^{out}$.

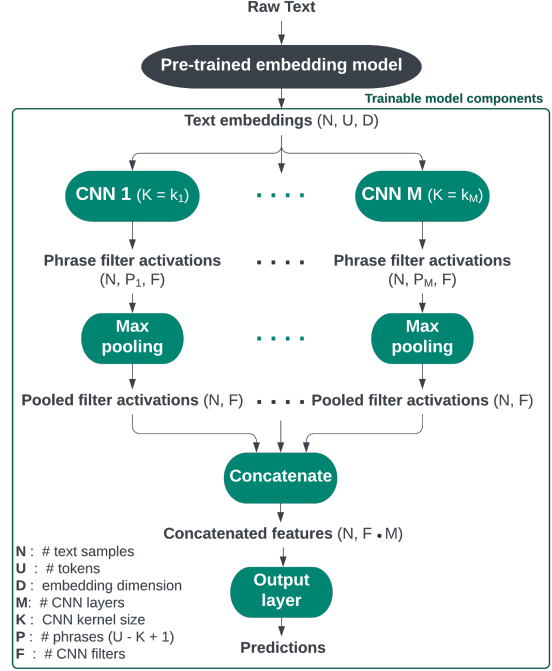

Figure 1: Model architecture

These penalty strengths and other hyperparameters are determined according to a fivefold cross validation procedure using the training set. Because the motivation of these models is primarily interpretation of learned features, rather than prediction performance, model selection is more subjective than simply choosing the highest accuracy parameter settings. We selected models based on a combination of accuracy, degree of correlation between filter activations (i.e. feature redundancy), and the number of "useful"[3] filters learned. Parameter settings for the models selected in our applications are reported in the appendix.

The final selected model is then re-trained using the entire training set with a randomly sampled 20% serving as the validation set, and is assessed using the unseen test set.

### 4.4 Identifying and testing influential text features

To interpret the model's learned latent representations and discover text treatments ($Z_i$) for each text, we utilize three model components:

1. The filter activations of each text sample's phrases for each filter $f$ ($a_{i,f}$);
2. The output layer weights ($W^{out} \in \mathbb{R}^{F \cdot M}$);
3. The input text samples ($T_i$).

The filter activations represent how strongly each phrase corresponds with the representation learned by each filter. To facilitate interpretation and to assign manual labels to each filter, we examine the phrases that maximally activate each filter. The final layer weights determine how each text representation contributes to the ultimate outcome prediction. Finally, the original input text samples provide context for the phrases that activate highly on each filter. This last component is most subjective to interpretation. Because input text embeddings are context-dependent, each phrases' embeddings contain more information than just the tokens that make up the phrase, which lack the context of the whole sample. However, due to the difficulty of interpreting text embedding dimensions, the context that human readers assign to phrases when reading an entire sample can not be confirmed to align with the encoded context.

---
[3]Some learned filter weights produce near-identical activations across samples. By not meaningfully distinguishing predictions between texts they are not useful for interpretation, so we omit filters with activation ranges less than a threshold $t = 0.05$ wide.

The objective of this interpretation process depends on whether the researcher wishes to directly estimate the effects of the identified latent features in the test set under assumptions described in Section 3, or if they wish to discover concrete text features to test in a follow up experiment. In the first scenario, the max-pooled filter activations ($a_{i,f}^{pooled}$) may be considered directly as the sample-level latent text treatments ($Z_i$). Researchers could also choose to binarize these features, for example by defining $Z_{i,f} = \mathbf{1}[a_{i,f}^{pooled} > \bar{a}_f^{pooled}]$ where $\bar{a}_f^{pooled}$ is the median of ($a_{i,f}^{pooled}$). This avoids the more stringent modeling assumptions needed for estimating effects of continuous treatments, though it may complicate interpretation. In either case, this interpretation process provides the researcher an understanding for what the latent text treatments represent and therefore the effects that they are estimating. In the second scenario, this process can guide the researcher's process for obtaining concrete text treatments. Here, a second set of text treatments, $\tilde{Z}_i$, are established which are not latent in the same sense as $Z_i$, because researchers control their definition. For example, $\tilde{Z}_i$ could be defined as an indicator of whether the manual labels assigned to a filter appear in experimental texts, or as measures of different tones or grammatical structure identified by filters.

### 4.5 Evaluation methods

We evaluate our models by comparing them to two benchmark methods. The first is the methodology proposed in Fong and Grimmer (2016), which uses a topic modeling approach to discover and interpret latent text treatments. We abbreviate this method as F&G. The second is regularized logistic regression on the vocabulary of n-grams in the corpus, which we abbreviate as RLR. Methods are compared quantitatively by assessing the adjusted R-squared of linear models fit using the text treatments identified by each method to predict the outcome variable, and by assessing the mean-squared error of these linear models on out-of-sample texts. We make these comparisons robust to sampling variability by calculating the metrics across 1000 bootstrap samples of the data, fixing the trained mod-els (and therefore the learned latent features). To better understand the stability of the training process for our proposed model and benchmark models, we repeat the process but additionally retrain the models (fixing the tuned parameter settings) in 150 bootstrap samples of the training data. Methods are compared qualitatively by assessing the interpretability and variety of learned text features. In the censorship application, ground-truth information of which phrases led to censorship allow us to compare methods by their ability to recover text treatments which have known causal effects on the outcome. Details of how benchmark methods were implemented and full interpretation results are included in Appendix B.

| $f$ | $W_f^{out}$ | $\widehat{\beta}_f$ | CI | Top extracted phrases (translated) | Known censored phrase |
|---|---|---|---|---|---|
| 1 | 1.42 | 0.23 | [0.18, 0.28] | "[CLS]Wuhan Institute of Virology Party","Wuhan Institute of Virology Specialty","[CLS]Wuhan Institute of Virology'" | "Wuhan virus" |
| 2 | 1.32 | 0.23 | [0.19, 0.28] | "Profiting from national disasters, such people", "Chinese virus said that some people", "Profiting from national disasters, such as some people" | "Profiting from national disasters" |
| 3 | 1.21 | 0.26 | [0.21, 0.31] | "Secretary of the Provincial Party Committee of a province", "Chen Quanjiao of the Poison Institute stated", "Renowned Secretary of the Hubei Provincial Party Committee" | "Provincial party secretary" |
| 9 | 0.91 | 0.07 | [0.02, 0.11] | "Diagnosis and Shincheonji Teaching", "Always waiting for Shincheonji Teaching" | "Shincheonji Church" |
| 10 | 0.77 | 0.11 | [0.07, 0.16] | "Jiang Chaoliang is in Wuhan" | "Jiang Chaoliang" |

Table 1: Frequent censorship rationale is learned by the model. The first column identifies filters in order of the weight $W_f^{out}$ assigned to their max-pooled activations $a_{i,f}^{pooled}$ in the final model layer (second column). The third column shows the average estimated treatment effect of $a_{i,f}^{pooled}$ across 1000 bootstrap samples of the test set (fixing the model and therefore the filters), with corresponding bootstrap confidence intervals in the fourth column. The fifth column lists filters' unique top 3 most associated phrases from the test set. The sixth column associates each filter with a commonly reported censored phrase.

## 5 Experiments

In order to sufficiently demonstrate the qualitative and substantive results that are achievable with our method, we focus our experiments on two datasets. The first was selected because of rare access to ground-truth rationales underlying the text-driven outcome. The second was chosen to explore a setting where influential text features likely exhibit complex and varied structures and because of its use in closely related research. We take a depth versus breadth approach here, but future work should assess this method on a larger range of datasets. While the method can generalize to any dataset where text is thought to cause an outcome, datasets with ground-truth information about this relationship are ideal (though rare) for demonstrating successful identification of causal relationships and effects.

### 5.1 Weibo post censorship

**Dataset and setup** For our first application, we use a sample of 28,386 Weibo posts from the Weibo-Cov dataset (Hu et al., 2020). These are social media posts on the topic of COVID and were posted in February 2020 on Weibo.[4] To obtain the censorship label for each post, we use the content review API from Baidu.[5] The API is a classifier that returns the probability of censorship for each post. The API only returns a probability of 1 when a social media post includes words or phrases that are on Baidu's blacklist. As the API also returns the flagged keywords and phrases, this enables us to validate whether our model can recover keywords and phrases that led to censorship.

We train our model to predict whether or not a post was flagged by the API to be cen-sored with probability 1. Although this outcome is not determined by direct human decision making, we can view the blacklist as a decision maker that perfectly implements a set of human-defined preferences (which may or may not be generally representative of the broader censorship policy). To tokenize and embed these texts, we use a pre-trained BERT Chinese language model provided by the Joint Laboratory of HIT and iFLYTEK Research, `MiniRBT-h288` (Yao et al., 2023).[6] This model has an embedding dimension of 288 and 12.3M parameters. The embeddings from the BERT model's last hidden state are used as the input features to our model architecture. Examples of posts in this data set, their censor probabilities, and their censor words (when applicable) with English translations are shown in Appendix A Table 3. Appendix A Table 4 shows the top 10 censor words across all censor-probability-one samples, their translations, and the proportion of censored samples corresponding to each.

---
[4]The data set creators anonymized identifiable information in posts to protect user privacy. Data is available from the creators upon request.
[5]ai.baidu.com/solution/censoring

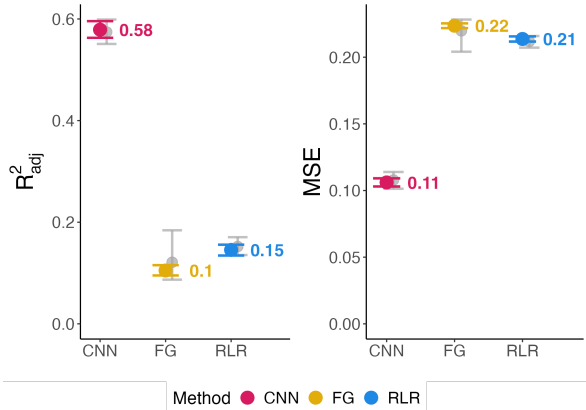

Figure 2: Model fit metrics from linear regression of censorship against text treatments identified by the CNN-based model and benchmarks across 1000 bootstrap samples with fixed trained model. Means and 2.5% and 97.5% quantiles of bootstrap distributions are shown. Gray marks show results when models are additionally retrained on resampled training data to assess overall stability.

**Results** The trained model obtains an accuracy score of 0.87 on the test set. Across iterations where the model is repeatedly trained on resampled data and reevaluated, the model classification accuracy is relatively stable (with values in [0.84, 0.88] and an average of 0.86). This performance indicates that the model has learned useful representations of Weibo posts from the time period which are predictive of censorship. Based on parameter tuning results, this model was constructed with two convolutional layers with kernel sizes set to 5 and 7 Chinese characters. We highlight our interpretation of the most relevant representations in Table 1, with interpretation of all representations included in Appendix A Table 7. We find that the two most commonly censored phrases, "Wuhan virus" (23.9% of censored posts) and "national crisis" (4.9% of censored posts) are clearly identified by the model in the first and second model filters. The max-pooled activations for these filters also contribute the most to the model's final prediction of censorship, as seen in the $W_f^{out}$ column of this table. The most highly-activating phrases for filters 3 and 9 have in common two other known censored phrases, "Provincial party secretary" and "Shincheonji Church," and the highest activated phrases for filter 10 concentrate exactly around the same phrase, which relates to a fifth known censor phrase "Jiang Chaoliang." The complete set of representation interpretations demonstrates that there is some redundancy in the keywords learned by filters. Their differences in sentence structure and context could be illuminating in other settings, though in this case it is known that the inclusion of these phrases solely affects the outcome. As a proof-of-concept, we include the effect estimates obtained by regressing the labels on the max-pooled filter activations of the test sample, though this setting does not follow a typical experimental design. Though the magnitude of the estimated effects differ from the output layer weights (in large part because the output layer weights correspond to a sigmoid rather than linear activation), they are in relative agreement about which text treatments are found to be most influential for censorship.[7]

**Model validation** We find that this methodology successfully recovers the phrases which cause the most posts to be censored. In a setting without oracle knowledge of the censorship rationale, we feel confident that researchers would be able to use this model to determine at least five of the most common censored phrases. In comparison, we find that the topics learned by the F&G method are not clearly aligned with any of the most common censored phrases (Appendix B: Table 12). A subset of the n-grams selected by the logistic regression model correspond to three common censored phrases: "Wuhan virus", "Jiang Chaoliang" and "Provincial party secretary" (Appendix B: Table 13). Our model outperforms both benchmarks with respect to both metrics reported in Figure 2, which indicates that the features learned by our model explain significantly more variation in the censorship outcome and have much better predictive power compared to the topics and keywords learned by the benchmark methods.

---

[6] Model is licensed under Apache License 2.0.

[7] The regression estimates $\beta_f$ identify causal effects under the assumptions in Section 3. Because of the much more complex way that the final layer weights $W_f^{out}$ are learned in the neural network model, we have not proven that they identify causal effects even under the same assumptions.

| $f$ | $W_f^{out}$ | $\widehat{\beta}_f$ | CI | Top extracted phrases | Inferred Concept |
|---|---|---|---|---|---|
| 1 | 0.77 | 0.01 | $[-0.09, 0.10]$ | "they were entered as if","inquiries were conducted on my","card were stolen . the","they were rejected , we","items are blended into my" | credit disputes |
| 2 | 0.77 | 0.15 | $[0.05, 0.25]$ | "to non - renew its","to place a longer fraud","contract by post dating inadequate","to know if coa meric","to question or challenge their" | disputed action / infinitive verb |
| 3 | 0.75 | 0.32 | $[0.22, 0.43]$ | "direct deposit our accounts are","have made many er rone","have used our bank '","jersey ( we pre -","when processing our new mortgage" | banking processes |
| 12 | −0.72 | −0.09 | $[-0.20, 0.02]$ | "tell me so i refused","tell me that i owed","bills they say i owe","know what services i provide","es pe ct ful manner" | debt collection attempts |
| 13 | −0.73 | −0.20 | $[-0.31, -0.09]$ | "a der oga tory collection","a company called " west","" der oga tory remark","received a call from premium","[CLS] i get a voice" | reference to phone calls/voicemails |
| 14 | −0.84 | −0.05 | $[-0.16, 0.07]$ | "finance corp showed up with","un hel pf ul .","recovery group called me on","debt collector has been calling","police department like they were" | reference to previous interactions |
| 15 | −0.86 | −0.09 | $[-0.20, 0.02]$ | "' ve asked multiple times","' m committing fraud and","' t able to receive","' re talking about .","' t received any legal" | use of contractions |
| 16 | −0.94 | −0.10 | $[-0.22, 0.01]$ | "a pay day loan company","a pay day loan during","a pay day loan from","a pay day loan from","a pay day loan that" | payday loans |

Table 2: CFPB model interpretation for top 8 filters. Columns 1-5 correspond to those in Table 1. The sixth column contains a manual interpretation of the top extracted phrases.

## 5.2 Consumer Financial Protection Bureau complaint response

**Dataset and setup** For our second application, we use a dataset from Egami et al. (2018) of 54,816 consumer complaint narratives submitted to the Consumer Financial Protection Bureau (CFPB) from March of 2015 to February of 2016.[8] The outcome variable indicates whether or not the complainant received a timely response from the company filed against. Due to severe imbalance in the outcome variable, we proceed with a subsample of complaints which received a timely response (5136 timely and 1712 non-timely responses) combined with a class-weighted loss function. To tokenize and embed the complaint texts, we use a pre-trained BERT English language model bert-tiny trained by Google Research (Turc et al., 2019; Bhargava et al., 2021).[9] This model has an embedding dimension of 128 and 4M parameters.

**Results** The trained model obtains an accuracy score of 0.76 and an F1 Score of 0.33 on the test set. Across iterations where the model is trained on resampled data and reevaluated, the model classification accuracy is relatively consistent with values in $[0.72, 0.79]$ and an average of 0.75. F1 scores are observed to be more variable with values in $[0.12, 0.44]$ and an average of 0.31. Given the limited size of the data set used, the class imbalance, and the relative complexity of this learning task, it is unsurprising that this model achieves a lower performance, however, the learned representations still provide meaningful insight into the textual treatments in the dataset.

---

[8]Data is publicly available: https://www.consumerfinance.gov/data-research/consumer-complaints/. The CFPB removes personal information from complaints.
[9]Model is licensed under Apache License 2.0.

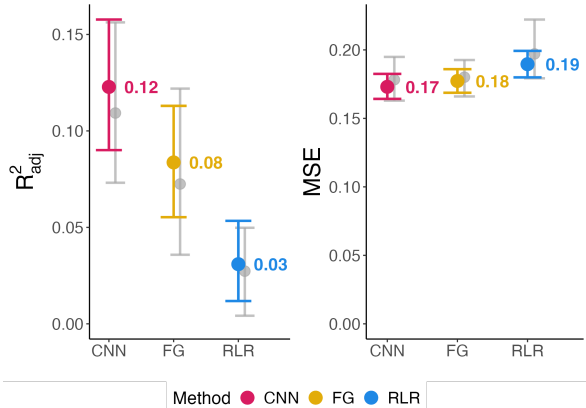

Figure 3: Model fit metrics from linear regression of timely response against text treatments identified by the CNN-based model and benchmarks across 1000 bootstrap samples with fixed trained model. Means and 2.5% and 97.5% quantiles of bootstrap distributions are shown. Gray marks show results when models are retrained on resampled training data to assess overall stability.

Table 2 summarizes interpretation of the top 8 representations (according to final layer weight) learned by this model. Interpretation for all filters can be found in Appendix A Table 10. This model has a single convolutional layer with kernel size set to 5 tokens, which was chosen by parameter tuning. We infer that references to credit disputes and banking processes may be positively associated with timely responses, and that references to attempted debt collection, phone calls or voicemails, previous interactions, or payday loans may be negatively associated with timely responses. Beyond these broader topics, we also find that using infinitive verbs in the context of describing disputed actions could increase the likelihood of timely response, while using contractions may have the opposite effect. Table 2 also includes effect estimates from regressing the test set labels against the texts' corresponding max-pooled filter activations. Again, we believe it is unlikely that the assumptions necessary for causal interpretation of these effects are met, but the estimates could still act as a useful tool for a researcher exploring possible text treatments for subsequent experiments.

**Model evaluation** In this application we do not have access to the true reasons that complaints receive or do not receive timely responses, and imagine that a variety of text features could impact this outcome. Both benchmark methods detected that certain financial topics seem to be associated with timely responses (Appendix B: Tables 15, 16). In particular, the results from all models imply that mentions of debt collection have a negative association, while mentions of banking processes and credit issues have a positive association, though the magnitude of estimates vary. Beyond these shared topics, our model is uniquely able to learn grammatical text features that are predictive of the outcome. While all models perform similarly in terms of the out-of-sample predictive power of their identified text features, our model results in a modestly higher $R^2_{adj}$ value compared to benchmarks (Figure 3).

## 6 Conclusion

We present a new method to discover influential text features represented by clusters of phrases of flexible length. Our approach is inspired by and builds upon previous work in computational social science and interpretable NLP, and provides experimenters with a quantitative tool for identifying promising text treatments to test in follow up experiments. When researchers are willing to make stronger identification assumptions discussed in Section 3, text treatments identified by using the model can also be used to estimate causal effects on the test set directly. Our applications demonstrate the ability of our model to learn useful and diverse latent text representations and its capacity to recover known influential text features. Important directions for future work include a human interpretability evaluation to rigorously compare the qualitative aspects of our method to benchmarks. Approaches for evaluating the stability of identified latent text treatments across model runs would also be useful to practitioners performing follow-up controlled experiments.

# 7 Limitations

**Small BERT models used out-of-the-box** In this paper, we do not investigate how model performance could be affected by fine-tuning the pre-trained BERT models, or by using larger or more advanced models to obtain higher dimensional word embeddings. Future work investigating how benefits from these changes trade-off with reduced computational efficiency would be relevant to researchers using this method.

**Computational expense** While these results demonstrate that the proposed CNN-based method is able to significantly outperform benchmark methods, these improvements come with a substantially higher computational cost in terms of both time and memory. We observed that the regularized logistic regression method was by far the most efficient option overall, with the F&G method also exhibiting much higher memory-efficiency but less time-efficiency. Future work comparing the computational expense of these methods across a larger range of datasets would be valuable for practitioners.

**Trade-off between experimental costs and less-interpretable treatments** Under the assumptions discussed in Section 3, researchers may estimate causal effects by directly testing the identified latent text treatments. This simplifies the experimental pipeline, but as in Egami et al. (2018) and Fong and Grimmer (2016), comes with the drawback of requiring the researcher to somewhat subjectively interpret the identified latent text treatments that are being tested. Alternatively, researchers may use their interpretations of the discovered latent text features to inspire "manifest" text treatments (ex. specific keywords, sentence structures) to test in confirmatory settings. In this case, the text features being tested would be known and manipulated by the researcher, allowing clearer interpretation of effects and weaker assumptions. This alternative comes with the downside of requiring researchers to run follow-up experiments.

**Incorporating uncertainty in latent treatments** Our paper does not provide guidance for incorporating the model-based uncertainty involved in identifying and estimating the latent text treatments into causal effect estimates.

**Designing experimental texts** We generally recommend using our model to guide the selection of text treatments for use in follow-up experiments. Designing experimental texts to isolate treatments of interest is a non-trivial task, and is left to the experimenter. In many cases, it is challenging to imagine altering a specific part of a text without affecting the interpretation of surrounding text that is not directly manipulated. This makes it difficult to establish causality for a specific text feature, rather than for the aggregate differences between a set of texts. This is a known challenge of making causal inferences with text, and relates to the strong ignorability assumption discussed in Section 3.

# 8 Ethics Statement

For any model designed to extract persuasive concepts, there is a risk that bad actors could use it to improve their ability to manipulate others. Many other tools exist which could presumably be used for this purpose, so we believe that the benefits of having this model open source outweigh this risk. An example of this kind of trade-off can be seen in the context of the model's application to censorship. When governments utilize human censors, they could potentially use this model to identify new keywords to add to an automated censorship blacklist to improve efficiency. On the other hand, the model can also be used to reverse engineer the process and reveal censorship policies, as we demonstrate. Acknowledging the possibility for misuse, we believe that the opportunities for productive and socially beneficial application are greater.

## Acknowledgements

This work was supported in part by the National Science Foundation RIDIR program, award numbers 1738411. We thank Kevin Li for contributions to an early version of this work. We thank the assigned anonymous ACL reviewers for their helpful feedback.

This is the author's version of work that has been accepted to Findings of ACL 2024 (https://doi.org/10.18653/v1/2024.findings-acl.714).

# A  Supplemental Results

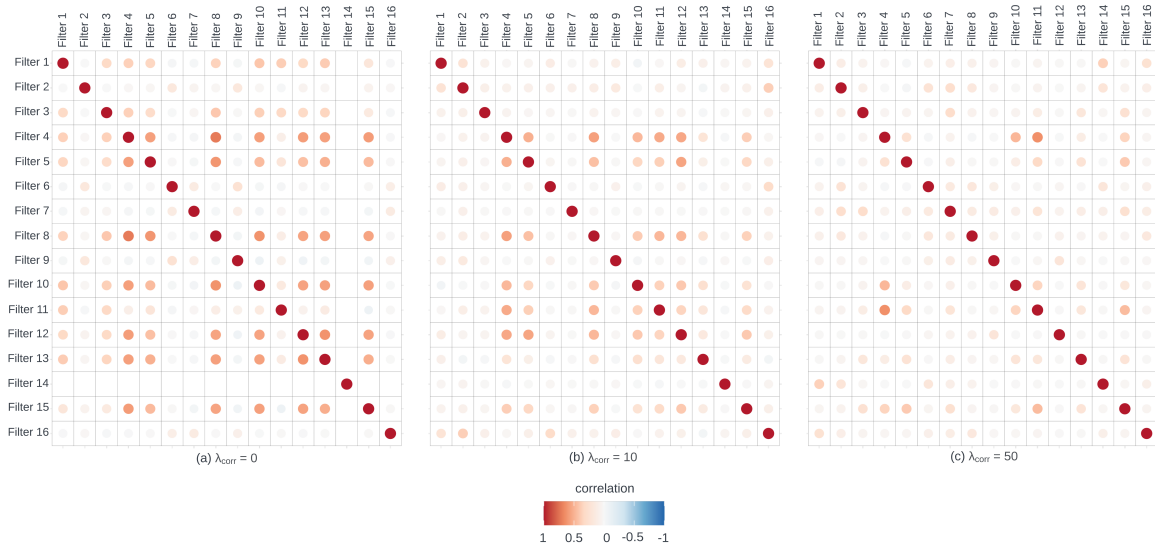

Figure 4: Correlation grids for filter activations when the correlation penalty is increased from (a) 0 to (b) 10 to (c) 50 for the censorship model. Dark red indicates pairwise correlation is closer to 1, dark blue indicates pairwise correlation is closer to -1, and white indicates pairwise correlation close to 0.

Example posts from the Weibo censorship data set

| Weibo post [translation] | Censorship probability | Censor keywords |
| --- | --- | --- |
| 武汉病毒所致信全所职工和研究生一首小诗，童年是一道彩虹，童年是一缕阳光。我把色我的童年印在一张张照片上，陪伴着我快乐地成长。[A letter from the Wuhan Institute of Virology to all employees and graduate students of the Institute: A little poem, childhood is a rainbow, childhood is a ray of sunshine. I printed my childhood on a photo and grew up happily with me.] | 1.0 | 武汉病毒 [Wuhan virus] |
| 疫情当前大发国难财，所售口罩均为三无产品怒怒怒说发货没有快递单号，退款均需扣费，请尽快查处怒怒怒 (tagged usernames omitted) [In the current epidemic situation, there is profiteering at the expense of the nation. All the masks sold are substandard products. Anger, anger, anger! It is claimed that shipments are made without providing a tracking number, and refunds will be subject to charges. Please investigate and resolve this issue as soon as possible. (tagged usernames omitted)] | 1.0 | 国难财 [Profiting from national disasters] |
| 点赞遵义遵义：一手抓防控一手抓经济，遵义复工复产全面铺开一手抓防控一手抓经济，遵义复工复产全面铺开转发理由: 转发微博 [Thumbs up for Zunyi. Zunyi: One hand focuses on epidemic prevention and control, and the other hand promotes economic development. Zunyi has comprehensively resumed work and production. Thumbs up for Zunyi. One hand focuses on epidemic prevention and control, and the other hand promotes economic development. Zunyi has comprehensively resumed work and production. Reason for reposting: Reposting Weibo.] | 0.5 | |
| 韩红捐赠的救援车进入雷神山韩红爱心慈善基金会捐赠的救护车进入雷神山了，整整齐齐的一排，谢谢韩红老师以及捐款的人嗷！！转发理由: good good good [The rescue vehicle donated by Han Hong entered Leishen Mountain. The ambulances donated by Han Hong Charity Foundation entered Leishen Mountain. They were lined up neatly. Thank you, Teacher Han Hong and those who donated！！Reason for forwarding: good good good] | 0.0 | |

Table 3: Sample posts from the Weibo post censorship data set. The first column contains sample posts and their translations into English. The second column is the probability of censorship, and the third column contains associated censorship keywords (when applicable) as returned by the Baidu API.

Most common censor keywords

| Censor keywords | Translation | % |
|---|---|---|
| 武汉病毒 | Wuhan virus | 23.9 |
| 国难财 | Profiting from national disasters | 4.9 |
| 抗肺炎 | Anti-pneumonia | 3.7 |
| 副省长 | Deputy Governor | 3.6 |
| 安倍晋三 | Shinzo Abe | 3.5 |
| 蒋超良-省委书记 | Jiang Chaoliang-Secretary of the Provincial Party Committee | 2.7 |
| 不作为 & 当地政府 | Inaction & local government | 2.4 |
| 省委书记 | Provincial party secretary | 2.3 |
| 省长 | Governor | 1.9 |
| 新天地教会 | Shincheonji Church | 1.9 |

Table 4: The 10 most common censor keywords in the Weibo post censorship data set. The first two columns contain words and phrases on Baidu's blacklist of censor keywords and their translations. The third column contains the percentage of justifications corresponding to each censor word/phrase.

| Hyper-parameter | Value |
|---|---|
| Number of tokens per sample | 150 |
| Number of filters per convolutional layer | 8 |
| Kernel sizes of conv. layers | 5, 7 |
| Conv. layer kernel regularizer penalty | 0.001 |
| Conv. layer activity regularizer penalty | 3 |
| Output layer kernel regularizer penalty | 0.0001 |
| Learning rate | 0.0001 |

Table 5: Hyper-parameter settings for the censorship model used to produce the reported results. This model has 27 681 trainable parameters total. During parameter tuning and the final model training, all models were trained for 100 epochs with early stopping (patience = 15) and batch sizes of 32.

| Tuned hyper-parameter | Values considered in tuning |
|---|---|
| Number of filters per conv. layer* | 4, 8, 16 |
| Kernel sizes of conv. layers | 5, 7, 5 and 7 |
| Conv. layer kernel regularizer penalty | 0, 0.0001, 0.001 |
| Conv. layer activity regularizer penalty | 0, 1, 3 |
| Output layer kernel regularizer penalty | 0.0001, 0.001, 0.01 |
| Learning rate | 0.00001, 0.0001, 0.001 |

Table 6: The censorship model parameter tuning process searched models with combinations of the above hyper-parameter values. Each model utilized 9.3 minutes of CPU time on average during training. The tuning procedure considered 486 different parameter settings, and with 5-fold cross validation for each setting utilized a total of 375 CPU hours across 4 cores. Each core was allocated 50GB of memory. Tuning was performed on a shared-resource computing cluster associated with our institution. *Models were required to have 8 or 16 total filters across convolutional layers. Combinations with one convolutional layer with 4 features, and models with two convolutional layers with 16 features each, were omitted from the tuning procedure.

Interpretation of all learned filters by the censorship model

| $f$ | $W_f^{out}$ | $\widehat{\beta}_f$ | Confidence Interval of $\widehat{\beta}_f$ | Top extracted phrases (translated) | Known censored phrases |
|---|---|---|---|---|---|
| 1 | 1.42 | 0.23 | [0.18, 0.28] | "[CLS] 武汉病毒所党","验武汉病毒所专","[CLS] 武汉病毒所'","? 武汉病毒所辟","。武汉病毒所所" ["[CLS]Wuhan Institute of Virology Party","Wuhan Institute of Virology Specialty","[CLS]Wuhan Institute of Virology'","? Created by the Wuhan virus", ". Wuhan Institute of Virology"] | "Wuhan virus" |
| 2 | 1.32 | 0.23 | [0.19, 0.28] | "国难财", 如此人","汉病毒所说某中","国难财比如某些","国难财也敢发，","国难财, 有些人" ["Profiting from national disasters, such people", "Chinese virus said that some people", "Profiting from national disasters, such as some people", "Profiting from national disasters, some people dare to make money,","Profiting from national disasters, some people"] | "Profiting from national disasters" |
| 3 | 1.21 | 0.26 | [0.21, 0.31] | "个省的省委书记","毒所陈全姣声明","任湖北省委书记","毒所的 remdesi" ["Secretary of the Provincial Party Committee of a province", " Chen Quanjiao of the Poison Institute stated", " Renowned Secretary of the Hubei Provincial Party Committee", " Remdesi of the Poison Institute."] | "Provincial party secretary" |
| 4 | 1.20 | 0.12 | [0.07, 0.17] | "病毒所党委","病毒所所长","病毒所研究","病毒所联合" ["Party Committee of the Institute of Virology", "Director of the Institute of Virology", "Research of the Institute of Virology", "Union of the Institute of Virology"] | "Wuhan virus" (using context of phrases within samples) |
| 5 | 1.18 | 0.05 | [0.00, 0.10] | "病毒所回应 6 大","病毒所所长已经","病毒所所长"（正）["The top 6 responses from the Institute of Virology", "Director of the Institute of Virology has been", "Director of the Institute of Virology" (positive)] | "Wuhan virus" |
| 6 | 1.09 | 0.10 | [0.06, 0.15] | "那些发国难","上是发国难","授旗。省委","期间发国难","任湖北省委" ["Those who caused national calamity", "The one who caused national calamity", "granted the flag. Provincial Party Committee", "During the national crisis", "Served as Hubei Provincial Party Committee"] | "National crisis" |
| 7 | 1.07 | 0.10 | [0.05, 0.15] | "武汉病毒所" ["Wuhan Institute of Virology"] | "Wuhan virus" |
| 8 | 1.03 | 0.15 | [0.11, 0.19] | "发国难财！","发国难财" ["Profiting from national disasters！","Profiting from national disasters"] | |
| 9 | 0.91 | 0.07 | [0.02, 0.11] | "确诊与新天地教","一直等新天地教","不保证打款时间" ["Diagnosis and Shincheonji Teaching", "Always waiting for Shincheonji Teaching", "No guarantee of payment time"] | "Shincheonji Church" |
| 10 | 0.77 | 0.11 | [0.07, 0.16] | "蒋超良在武" ["Jiang Chaoliang is in Wuhan"] | "Jiang Chaoliang" |
| 11* | −0.38 | - | - | "2020 我们需要的是",": 辛苦啦, 希望","! 辛苦了! 抱抱",", 东西都来不及","? 有坚持有希望" ["What we need in 2020 is",":Thank you for your hard work, hope","! Thanks for your hard work! Hug",", it's too late for anything","? "Persistence and hope"] | |

| | | | | |
|---|---|---|---|---|
| 12* | −0.48 | - | - | "购买防护及消毒","武汉加油！转发","铁、公交等公共","距离接触等条件","交往增多，临省" ["Purchase protection and disinfection", "Come on Wuhan! Forward", "Railway, bus and other public places", "Distance contact and other conditions", "Increased exchanges, close to the province"] |
| 13* | −0.66 | - | - | "战疫，我们","疫情，我们" ["Fight the epidemic, we", "Fight the epidemic, we"] |
| 14* | −0.80 | - | - | "上报的防疫","召开的疫情","条件的传染","其来的疫情" ["Reported epidemic prevention", "Convened epidemic", "Conditional infection", "Occurring epidemic"] |
| 15 | −1.06 | −0.08 | [−0.13, −0.03] | "国加油！心","国加油！加","子里凉凉了","[CLS]春暖花开","待春暖花开" ["Come on country! Heart", "Come on country! Add", "It's getting cold inside", "[CLS] The flowers are blooming in the spring", "Waiting for the flowers to bloom in the spring"] |
| 16 | −1.21 | −0.04 | [−0.09, 0.01] | "leban 乐班营业","今天是疫情开工","机器。泪泪家里","今天, 20200202,","过去, 老伙伴们" ["leban Leban is open for business", "Today is the start of the epidemic", "Machine. Tears at home","Today, 20200202,","In the past, old friends"] |

Table 7: Full results of censorship model filter interpretation. The first column identifies filters in order of the weight assigned to their max-pooled filter activations $a_{i,f}^{pooled}$ in the final model layer (second column). The third column shows the average estimated treatment effect of $a_{i,f}^{pooled}$ across 1000 bootstrap samples of the test set, with corresponding bootstrap confidence intervals in the fourth column. The fifth column lists the unique phrases within the top 5 test set phrases that were most associated with each filter. The sixth column associates filters with one of the top 10 most commonly reported censor words in the data set (blank if none are applicable). *The associated max pooled filter activations had a range of less than 0.05, and therefore were omitted from interpretation and the regression to estimate $\beta$.

| Hyper-parameter | Value |
|---|---|
| Number of tokens per sample | 250 |
| Number of filters per convolutional layer | 16 |
| Kernel sizes of conv. layers | 5 |
| Conv. layer kernel regularizer penalty | 0 |
| Conv. layer activity regularizer penalty | 0.5 |
| Output layer kernel regularizer penalty | 0.001 |
| Learning rate | 0.001 |

Table 8: Hyper-parameter settings for the CFPB model used to produce the reported results. This model has 10 273 trainable parameters. During tuning and the final model training, all models were trained for 100 epochs with early stopping (patience = 15) and batch sizes of 32.

| Tuned hyper-parameter | Values considered in tuning |
| --- | --- |
| Number of filters per convolutional layer* | 4, 8, 16 |
| Kernel sizes of conv. layers | 5, 7, 5 and 7 |
| Conv. layer kernel regularizer penalty | 0, 0.0001, 0.001, 0.01 |
| Conv. layer activity regularizer penalty | 0, 0.5, 1, 3 |
| Output layer kernel regularizer penalty | 0.0001, 0.001, 0.01 |
| Learning rate | 0.00001, 0.0001, 0.001, 0.01, 0.1 |

Table 9: Combinations of the above hyper-parameter values were considered for tuning of the CFPB model. Records of exact computational resources used are no longer available. Based on those used to train the final model (2 minutes of CPU time), we estimate that the tuning procedure, which considered 1440 different parameter settings with 5-fold cross validation for each, would have utilized about 240 CPU hours across 3 cores each with 40GB of memory. Tuning was performed on a shared-resource computing cluster associated with our institution. *Models were required to have 4, 8 or 16 total filters across convolutional layers. Combinations producing a model with two convolutional layers with 16 features each were omitted from the tuning procedure.

Interpretation of all learned filters by the CFPB model

| $f$ | $W_f^{out}$ | $\widehat{\beta}_f$ | Confidence Interval of $\widehat{\beta}_f$ | Top extracted phrases | Inferred concept | CD plot |
|---|---|---|---|---|---|---|
| 1 | 0.77 | 0.01 | [−0.09, 0.10] | "they were entered as if","inquiries were conducted on my","card were stolen . the","they were rejected , we","items are blended into my" | Credit disputes | |
| 2 | 0.77 | 0.15 | [0.05, 0.25] | "to non - renew its","to place a longer fraud","contract by post dating inadequate","to know if coa meric","to question or challenge their" | disputed action / infinitive verb | |
| 3 | 0.75 | 0.32 | [0.22, 0.43] | "direct deposit our accounts are","have made many er rone","have used our bank '","jersey ( we pre -","when processing our new mortgage" | banking processes | |
| 4* | 0.46 | - | - | "owed .  in fact ,","the interest only model would","the money was added to","bp o was used ,","a " good faith '" | debt management | |
| 5* | 0.37 | - | - | "ve requested statements and /","is the 1 . 5","just bad people . [SEP]","to establish contact with xx","doing what i ' m" | attempts to communicate | |
| 6* | 0.35 | - | - | "added my card and tried",". fixed mortgage . only","( with date and address","issued by mail and i","30 days ago and forgot" | action oriented phrases | |
| 7 | 0.33 | 0.00 | [−0.11, 0.10] | "part b , sub section","several forms and af fi","although funds were de ducted","exemption s . sub par","these items from my report" | legal dispute | |
| 8* | 0.12 | - | - | "to pay the bill .","to restore our property .","not state the borrow ers","went to the chase back","to utilize the governments " " | infinitive verb | |
| 9 | −0.41 | −0.06 | [−0.17, 0.04] | "response and recently received a","any point , especially a","she thought i had committed","this case , the debt","they said there was an" | debt dispute | |
| 10 | −0.44 | −0.08 | [−0.20, 0.03] | "are seeking assistance to get","no very well treat me","am extremely worried that they","s quite obvious i '","have a legitimate reason for" | emotional / situational justification | |
| 11 | −0.57 | 0.01 | [−0.09, 0.12] | "was giving me monthly .","claiming he was law enforcement","not treating me fairly .","could " re - age","said it was done ." | fairness/authority dispute | |
| 12 | −0.72 | −0.09 | [−0.20, 0.02] | "tell me so i refused","tell me that i owed","bills they say i owe","know what services i provide","es pe ct ful manner" | debt collection attempts | |
| 13 | −0.73 | −0.20 | [−0.31, −0.09] | "a der oga tory collection","a company called " west","" der oga tory remark","received a call from premium","[CLS] i get a voice" | reference to phone calls/voicemails | |
| 14 | −0.84 | −0.05 | [−0.16, 0.07] | "finance corp showed up with","un hel pf ul .","recovery group called me on","debt collector has been calling","police department like they were" | reference to previous interactions | |
| 15 | −0.86 | −0.09 | [−0.20, 0.02] | "' ve asked multiple times","' m committing fraud and","' t able to receive","' re talking about .","' t received any legal" | use of contractions | |
| 16 | −0.94 | −0.10 | [−0.22, 0.01] | "a pay day loan company","a pay day loan during","a pay day loan from","a pay day loan that" | payday loans | |

Table 10: Summary of the CFPB model's learned representations. The first column identifies filters in order of the weight assigned to their max-pooled filter activations $a_{i,f}^{pooled}$ in the final model layer (second column). The third column shows the average estimated treatment effect of $a_{i,f}^{pooled}$ across 1000 bootstrap samples of the test set, with corresponding bootstrap confidence intervals in the fourth column. The fifth column lists the unique phrases within the top 5 test set phrases that were most associated with each filter. The sixth column contains a manual interpretation of the concept identified by the top phrases. The seventh column displays conditional density plots for the max-pooled filter activations corresponding to each filter. The x-axis of these plots represents the filter activation value. The y-axis indicates estimated probability of belonging to the positive class (dark gray color). *The associated max pooled filter activations had a range of less than 0.05, and therefore were omitted from interpretation and the regression to estimate $\beta$.

## B  Benchmarks

We evaluate the models developed for each application by comparing them to two benchmark methods.

1. F&G: The methodology presented in Fong and Grimmer (2016) is motivated by the same setting as this paper. Rather than neural networks with convolutional structures, they utilize a topic modeling approach via the supervised Indian Buffet process to discover latent text treatments. We implement their methodology to both of our applications using the `texteffect` R package produced by the same authors. For each application the input is a word-document matrix, with vocabulary restricted by excluding stop words and words that appear infrequently ($< 400$ times for the censorship data set and $< 300$ times for the CFPB data set). For the censorship application, we perform word segmentation using the `jiebaR` package. Using the training set, we perform a parameter search using the `sibp_param_search` function over `alphas` in $\{3, 5\}$ and `sigmasq.ns` in $\{0.2, 0.4, 0.6\}$. For both applications, we set the number of topics to $K = 16$ so that we obtain the same number of latent text treatments as our models identify. The final text treatments are chosen using the `sibp_rank_runs` function.

2. Regularized logistic regression (RLR): We perform regularized logistic regression with a L1 penalty on 3-grams in each corpus. This method is simple relative to the model presented in this paper and to the F&G model, but offers a very clear interpretation of the text features that are selected as predictors of the outcome. Instead of requiring subjective labeling of latent features, this method simply selects 3-grams whose in-sample frequencies are predictive of the outcome. We chose 3-grams to ensure that phrase lengths would be similar to those highlighted in our model, as BERT uses sub-word tokenization. For both applications we exclude stop words and only consider n-grams with frequencies over a certain threshold to control vocabulary size (n-grams with frequencies $< 50$ were excluded from the CFPB analysis, and $< 200$ from the censorship analysis). We chose the penalty parameter to be the minimum magnitude such that at most 16 3-grams were selected by each model. For the censorship data, multiple selected 3-grams were perfectly collinear with other 3-grams and were dropped from the final regressions and comparisons. This tuning and variable selection process was performed using the training set, and final estimates were computed using the test set.

We use two metrics to quantitatively assess the extent to which our model's learned latent text features predict the outcome compared to benchmark methods. First, we performed linear regressions of the outcome against the features identified by each model to compare Adjusted $R$-squared values. This assesses the degree to which latent features captured variation in the outcome. Feature representations are learned (F&G method) and n-grams are selected (regularized logistic regression method) in the training set, and evaluation of $R^2_{adj}$ is calculated using the the test set. Second, we calculated the prediction accuracy (mean-squared error) of regression models fit to each method's latent features in the training set when applied to the test set. These quantitative results are shown in the main body of the paper.

A key motivation for this work is to enable researchers to identify and understand text treatments from a text corpus. Therefore an important, but subjective, feature to compare between our models and benchmarks is the interpretability of identified text features. For this purpose, we list simplified versions of our convolutional filter interpretation tables, tables with the top keywords per topic and manual topic labels from the F&G method results, and tables with the selected n-grams from the regularized logistic regression method. In the censorship case, we can also compare the ability of the models to recover keywords that are known to cause censorship in the corpus.

Finally, the extent to which latent features correlate with each other across methods may also provide insights into the patterns that each method is successful or not successful at detecting.

To assess this, we include correlation plots for our method's identified latent features paired with those identified by each of the other methods.

## B.1 Censorship

**CNN model latent features**

| Manual label | Examples | Estimate |
|---|---|---|
| wuhan institute of virology, Wuhan virus | "CLSWuhan Institute of Virology Party","Wuhan Institute of Virology Specialty","CLSWuhan Institute of Virology'","? Created by the Wuhan virus", ". Wuhan Institute of Virology" | 0.22 |
| profiteering from national disasters | "Profiting from national disasters, such people", "Chinese virus said that some people", "Profiting from national disasters, such as some people", "Profiting from national disasters, some people dare to make money,","Profiting from national disasters, some people" | 0.24 |
| party secretary | "Secretary of the Provincial Party Committee of a province", " Chen Quanjiao of the Poison Institute stated", " Renowned Secretary of the Hubei Provincial Party Committee", " Remdesi of the Poison Institute. | 0.25 |
| wuhan institute of virology affiliation | "Party Committee of the Institute of Virology", "Director of the Institute of Virology", "Research of the Institute of Virology", "Union of the Institute of Virology" | 0.12 |
| wuhan institute of virology director | "The top 6 responses from the Institute of Virology", "Director of the Institute of Virology has been", "Director of the Institute of Virology" | 0.06 |
| national crisis | "Those who caused national calamity", "The one who caused national calamity", "granted the flag. Provincial Party Committee", "During the national crisis", "Served as Hubei Provincial Party Committee" | 0.11 |
| wuhan institute of virology | "Wuhan Institute of Virology" | 0.11 |
| profiteering from national disastors | "Profiting from national disasters! ","Profiting from national disasters" | 0.15 |
| shincheonji church | Diagnosis and Shincheonji Teaching, "Always waiting for Shincheonji Teaching", "No guarantee of payment time" | 0.07 |
| jiang chaoliang | Jiang Chaoliang is in Wuhan | 0.11 |
| support/gratitude | "What we need in 2020 is",":Thank you for your hard work, hope","! Thanks for your hard work! Hug",", it's too late for anything","? "Persistence and hope" | - |
| public health | "Purchase protection and disinfection", "Come on Wuhan! Forward", "Railway, bus and other public places", "Distance contact and other conditions", "Increased exchanges, close to the province" | - |
| resilience | Fight the epidemic, we, "Fight the epidemic, we" | - |
| epidemic dynamics | "Reported epidemic prevention", "Convened epidemic", "Conditional infection", "Occurring epidemic" | - |
| seasonal hope | "Come on country! Heart", "Come on country! Add", "It's getting cold inside", "CLS The flowers are blooming in the spring", "Waiting for the flowers to bloom in the spring" | -0.09 |
| pandemic onset | "leban Leban is open for business", "Today is the start of the epidemic", "Machine. Tears at home","Today, 20200202,","In the past, old friends" | -0.04 |

Table 11: CNN model latent text features. Estimates shown are from a single run and are blank for inactive filters.

**F&G model latent features**

| Manual label | Examples | Estimate |
| --- | --- | --- |
| tcm early findings | Early stage, Yes, Elevate, answer, through train, clinical medicine, kindness, ask, verify, too much | 0.32 |
| tcm early findings | Virus, drug, Research, Early stage, Shuanghuanglian, flu virus, Can be suppressed, Shanghai, Discover, at present | 0.32 |
| tcm guidance | Syndrome differentiation and treatment, doctor's orders, obey, Traditional Chinese Medicine, does not equal, break away, Don't, remind, use, specific | 0.30 |
| tcm modern research | Research, Oral liquid, Chinese patent medicine, Can be suppressed, Huazhong University of Science and Technology, Shuanghuanglian, The institute, learned, joint, Shanghai | 0.27 |
| tcm early findings | Early days, answer, Yes, ask, Elevate, kindness, verify, clinical medicine, Early stage, through train | 0.25 |
| public accountability | Step down, catharsis, get scolded, See, speak out, absolute, wide awake, exposed, forget, Apologize | 0.23 |
| tcm early findings | Early stage, efficient, Preliminary, Yes, answer, Elevate, ask, Shuanghuanglian, kindness, through train | 0.11 |
| patient zero | researcher, graduate School, Zero, Huang Yanling, beautiful, Chen Quan, postgraduate, ensure, Wuhan, Tie | 0.04 |
| public criticism | Step down, catharsis, get scolded, See, speak out, question, not good, exposed, forget, wide awake | 0.02 |
| tcm guidance | used for, obey, doctor's orders, Traditional Chinese Medicine, Syndrome differentiation and treatment, break away, remind, The medicine, Don't, does not equal | 0.01 |
| tcm guidance | does not equal, obey, doctor's orders, Traditional Chinese Medicine, Syndrome differentiation and treatment, specific, break away, use, Don't, The medicine | -0.03 |
| tcm early findings | Yes, answer, Elevate, clinical medicine, kindness, verify, ask, through train, Early days, benefit | -0.05 |
| public accountability | get scolded, catharsis, Step down, See, speak out, Apologize, exposed, forget, wide awake, absolute | -0.14 |
| public accountability | catharsis, Step down, get scolded, See, speak out, wide awake, forget, absolute, exposed, Apologize | -0.15 |
| tcm guidance | does not equal, obey, doctor's orders, Syndrome differentiation and treatment, specific, Traditional Chinese Medicine, break away, Don't, The medicine, remind | -0.44 |
| tcm treatment | Jianping, three flavors, Inside and outside, Shuangqing, new use, syndrome, broad spectrum anti, have, Detoxification, Jiang Hualiang | -1.21 |

Table 12: Fong & Grimmer model latent text features. TCM is an abbreviation for "Traditional Chinese Medicine."

**RLR model latent features**

| Label | Examples | Estimate |
|---|---|---|
| Do _a lot_ of experiments | - | 0.09 |
| Can suppress _new_ coronavirus | - | 0.08 |
| Hubei_Vice Governor_Response | - | 0.07 |
| See people all over the country | - | 0.06 |
| Hubei Provincial Party Committee_Secretary_Jiang Chaoliang | - | 0.06 |
| Wuhan_Virus_Institute | - | 0.04 |
| Chinese Academy of Sciences_Wuhan_Virus | - | 0.03 |
| Chinese Academy of Sciences_Shanghai_Drug | - | 0.03 |
| Indeed _exposed_ too | - | 0.00 |
| The epidemic is indeed exposed | - | 0.00 |
| Patient_valid_still | - | -0.04 |
| Weibo_Lottery_Platform | - | -0.04 |

Table 13: Selected features from the regularized logistic regression model on n-gram counts, where the penalty is minimized under the requirement that 16 or less features are selected (the number of filters in our CNN model). Here, features correspond simply to the count of specific n-grams in each text, so there is no distinction between the feature label and examples used to arrive at that label.

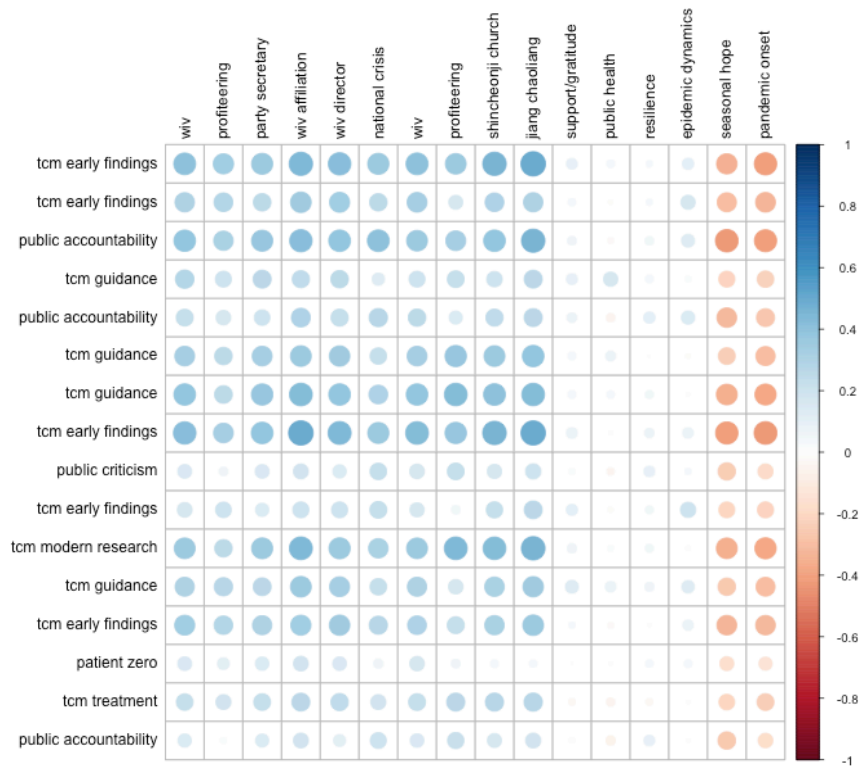

Figure 5: Correlation between the latent features learned by the CNN model (columns) and the latent features learned by the Fong & Grimmer method (rows).

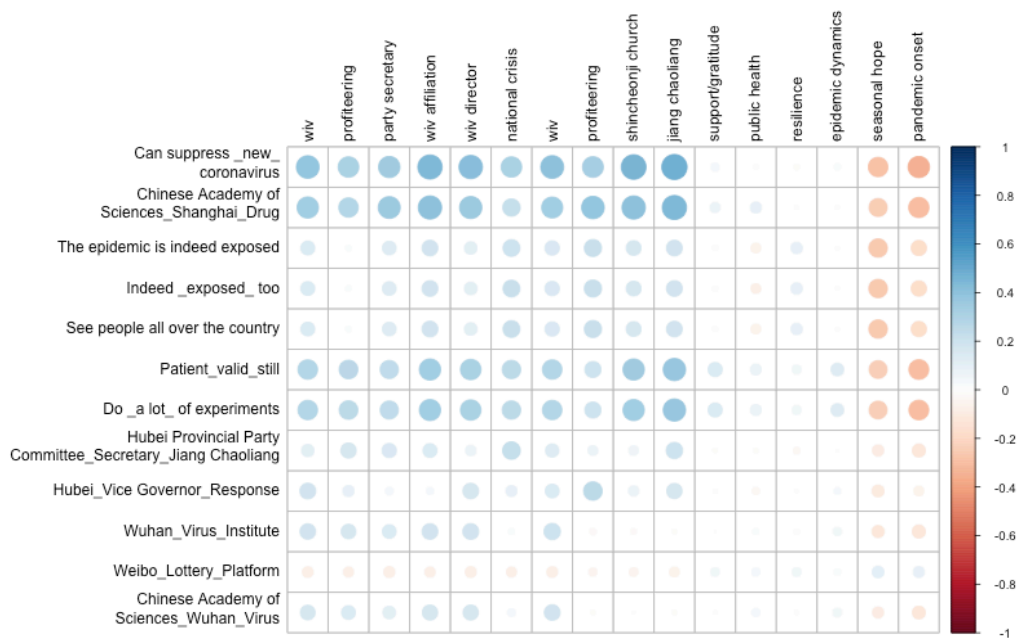

Figure 6: Correlation between the latent features learned by the CNN model (columns) and the latent features learned by the regularized logistic regression method (rows).

## B.2 CFPB

### CNN model latent features

| Manual label | Examples | Estimate |
|---|---|---|
| credit disputes | "they were entered as if,"inquiries were conducted on my,"card were stolen . the,"they were rejected , we,"items are blended into my | 0.01 |
| disputed action / infinitive verb | "to non - renew its,"to place a longer fraud,"contract by post dating inadequate,"to know if coa meric,"to question or challenge their | 0.15 |
| banking processes | "direct deposit our accounts are,"have made many er rone,"have used our bank ',"jersey ( we pre -,"when processing our new mortgage | 0.32 |
| debt management | "owed . in fact ,,"the interest only model would,"the money was added to,"bp o was used ,,"a " good faith ' | - |
| attempts to communicate | "ve requested statements and /,"is the 1 . 5,"just bad people . [SEP],"to establish contact with xx,"doing what i ' m | - |
| action oriented phrases | "added my card and tried,". fixed mortgage . only,"( with date and address,"issued by mail and i,"30 days ago and forgot | - |
| legal dispute | "part b , sub section,"several forms and af fi,"although funds were de ducted,"exemption s . sub par,"these items from my report | -0.01 |
| infinitive verb | "to pay the bill .,"to restore our property .,"not state the borrow ers,"went to the chase back,"to utilize the governments " | - |
| debt dispute | "response and recently received a,"any point , especially a,"she thought i had committed,"this case , the debt,"they said there was an | -0.07 |
| emotional/situational justification | "are seeking assistance to get,"no very well treat me,"am extremely worried that they,"s quite obvious i ',"have a legitimate reason for | -0.08 |
| fairness/authority dispute | "was giving me monthly .,"claiming he was law enforcement,"not treating me fairly .,"could " re - age,"said it was done . | 0.01 |
| debt collection attempts | "tell me so i refused,"tell me that i owed,"bills they say i owe,"know what services i provide,"es pe ct ful manner | -0.08 |
| reference to phone calls/voicemails | "a der oga tory collection,"a company called " west,""" der oga tory remark,"received a call from premium,"[CLS] i get a voice | -0.21 |
| reference to previous interactions | "finance corp showed up with,"un hel pf ul .,"recovery group called me on,"debt collector has been calling,"police department like they were | -0.05 |
| use of contractions | "' ve asked multiple times,"' m committing fraud and,"' t able to receive,"' re talking about .,"' t received any legal | -0.09 |
| payday loans | "a pay day loan company,"a pay day loan during,"a pay day loan from,"a pay day loan that | -0.10 |

Table 14: CNN model latent text features. Estimates shown are from a single run and are blank for inactive filters.

## F&G model latent features

| Manual label | Examples | Estimate |
|---|---|---|
| property financing and taxes | mortgage, home, loan, modification, foreclosure, property, sale, year, taxes, escrow | 0.14 |
| credit card services | account, customer, card, service, online, credit, checking, charged, charge, check | 0.10 |
| payment processing | payment, made, told, received, xxxx, payments, month, paid, pay, bank | 0.08 |
| credit reporting, FCRA | account, reporting, act, consumer, credit, fair, following, please, due, attached | 0.07 |
| home financing | modification, loan, mortgage, foreclosure, rate, income, home, sale, application, submitted | 0.07 |
| mortgage documentation | loan, xxxx, mortgage, 3, documents, complaint, 2, regarding, 1, modification | 0.06 |
| credit report dispute | report, credit, information, removed, verify, disputed, remove, dispute, reporting, asked | 0.06 |
| credit dispute | xxxx, act, filed, credit, due, original, made, provide, also, violation | 0.03 |
| employment issues | n't, get, help, job, got, know, work, going, just, now | 0.00 |
| late payments | calls, payments, calling, payment, month, monthly, stop, late, paying, call | -0.04 |
| debt validation | act, debt, fair, validation, collect, please, provide, violation, provided, consumer | -0.04 |
| bank communication | call, phone, calling, called, calls, number, back, branch, someone, told | -0.08 |
| communication issues | card, called, give, n't, call, calling, social, said, just, person | -0.10 |
| debt collection (legal) | debt, collection, creditor, fair, violation, validation, agency, collector, law, act | -0.11 |
| customer service | call, called, asked, said, told, message, left, back, weeks, sale | -0.16 |
| debt collection (logistical) | debt, collection, agency, reporting, report, creditor, collect, owe, credit, collections | -0.19 |

Table 15: Fong & Grimmer model latent text features.

## RLR model latent features

| Label | Examples | Estimate |
|---|---|---|
| credit_card_account | - | 0.02 |
| credit_reporting_agencies | - | 0.02 |
| bank_xxxx_xxxx | - | 0.00 |
| information_credit_report | - | 0.00 |
| attempting_collect_debt | - | 0.00 |
| call_xxxx_xxxx | - | 0.00 |
| fair_debt_collection | - | -0.01 |
| phone_xxxx_xxxx | - | -0.01 |
| attempt_collect_debt | - | -0.01 |
| named_xxxx_xxxx | - | -0.02 |
| xxxx_xxxx_stating | - | -0.03 |
| debt_collection_agency | - | -0.03 |
| phone_number_xxxx | - | -0.03 |
| debt_xxxx_xxxx | - | -0.04 |

Table 16: Regression model latent text features.

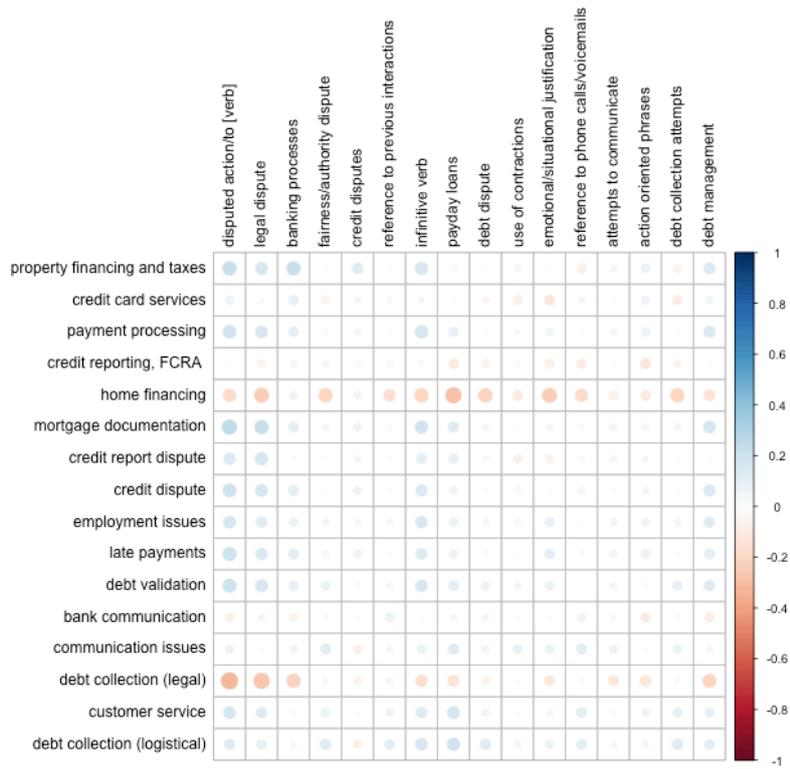

Figure 7: Correlation between the latent features learned by the CNN model (columns) and the latent features learned by the Fong & Grimmer method (rows).

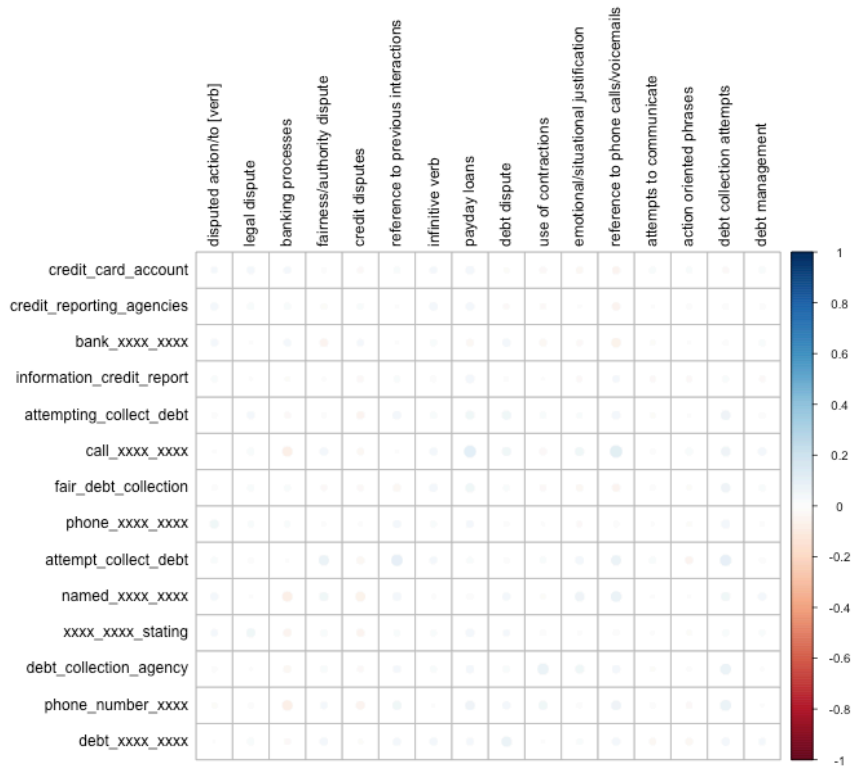

Figure 8: Correlation between the latent features learned by the CNN model (columns) and the latent features learned by the regularized logistic regression method (rows).